# Clustering Analysis of Interactive Learning Activities Based on Improved BIRCH Algorithm

Xiaona Xia[*]

*Faculty of Education, Qufu Normal University, Qufu, Shandong, 273165; School of Computer Science, Qufu Normal University, Rizhao, Shandong, 276826, China; Chinese Academy of Education Big Data, Qufu Normal University, Qufu, Shandong, 273165*

Group tendency is a research branch of computer assisted learning. The construction of good learning behavior is of great significance to learners' learning process and learning effect, and is the key basis of data-driven education decision-making. Clustering analysis is an effective method for the study of group tendency. Therefore, it is necessary to obtain the online learning behavior big data set of multi period and multi course, and describe the learning behavior as multi-dimensional learning interaction activities. First of all, on the basis of data initialization and standardization, we locate the classification conditions of data, realize the differentiation and integration of learning behavior, and form multiple subsets of data to be clustered; secondly, according to the topological relevance and dependence between learning interaction activities, we design an improved algorithm of BIRCH clustering based on random walking strategy, which realizes the retrieval evaluation and data of key learning interaction activities; Thirdly, through the calculation and comparison of several performance indexes, the improved algorithm has obvious advantages in learning interactive activity clustering, and the clustering process and results are feasible and reliable. The conclusion of this study can be used for reference and can be popularized. It has practical significance for the research of education big data and the practical application of learning analytics.

## 1. Introduction

Online technology and data technology have greatly promoted the developments of education big data (Christian Fischer et al., 2020; Dahdouh K et al., 2020) and intelligent education (Liu Min, & Zheng Mingyue, 2019; Chen Kaiquan, et al., 2019) , the integration and implementation of education and teaching process and data analysis technology make process tracking and learning analysis possible, which is the key application of data-driven education decision-making (Silva R et al., 2020). Through data mining, we can realize the proper description of education and teaching process, and through data analysis, we can have insight into the relevant problems in the process of education and teaching. These two aspects work together to achieve decision-making. Learning interaction activity is an important expression of learners' participation in the process of education and teaching. The topological relationships between activities form learning behaviors, which is the basis of realizing decision feedback. Clustering of learning interaction activities is an important part of topological relationship analysis, which can also serve personalized learning behavior research and individualized teaching practice (hosch B, 2020).

---

[*] Corresponding author. Email: xiaxn@qfnu.edu.cn

The data generated by online learning interaction activities is massive, discrete and autonomous, and the amount of information contained in the data is sparse. It is difficult to form a continuous and complete description of learning interaction activities, original data can not be minded to get useful value. It's very necessary to extract useful information to acquire knowledge or wisdom by mining the internal relationships of learning interaction activities. Learning interaction activities have always been a typical application of data analysis technology in education big data (Hu Yiling, & Gu Xiaoqing, 2019; Avila C et al., 2020; Saqr m et al., 2020).

There are two types of learning interaction: group topology and individual topology. For group topology, they reflect the group tendency (Li Yanyan et al., 2019) and individual preferences of learners respectively (Liu Min, & Zheng Mingyue, 2019), that needs to expand the cluster analysis of relevant data, find out the correlation and relationship strength from the distribution of clusters, and the clustering process and results determine the topological relationships of learning interaction activities. From the learners' learning behavior, we can extract the effective learning path, combine the learning contents and the features of learners, and play a positive role in quoting learners to gradually establish their own learning path. However, in general, learning interaction activities involve multi-attribute, multi feature and multi structure, and the relationships of activities are relatively complex, which creates a big obstacle to cluster analysis. There are few results focused on the cluster analysis of learning behavior, and there are many difficulties in the systematic comparative test (Garcia DIAS R et al., 2020; Jos é Maia Neto et al., 2020; Brisco N D A et al., 2020). At the same time, about learning interaction activities, the traditional clustering algorithm is not suitable for data processing and path retrieval, and is not conducive to the available and valuable results the analysis process has many limitations. Therefore, the feasible clustering of learning behavior needs to analyze the data structure and relationships, and design the corresponding data structure optimization algorithm.

Data clustering is a statistical analysis method for the known data or determined indicators. Based on the known features, the data are divided into different groups according to the similarity and dissimilarity, the groups are usually called clusters. The objects contained in the clusters generally have high similarity of certain features, while the objects contained in different clusters have obvious dissimilarity. Clustering analysis is a typical unsupervised learning process. Common clustering algorithms include K-means clustering (Liu S, & Zou Y, 2020), hierarchical clustering (Mulka M, & lorkiewicz W, 2020), probability density based scanning clustering (Yang Tianpeng, & Chen Lifei, 2018) and Gaussian clustering model (Zhao Y H et al., 2020); Tao Zhiyong et al., 2018), but these algorithms have their own advantages and disadvantages, which are not suitable for all data. For example, K-means clustering measures data features in the same proportion, so it is necessary to know the exact number of data sets in advance; hierarchical clustering improves the disadvantage of K-means, which is significantly sensitive to outliers. It can retrieve data independently and realize hierarchical clustering in the form of tree graph, but the calculation process of the algorithm is complex, and the software and hardware environment supporting the algorithm is highly required. The scanning clustering algorithm based on probability density also does not need the exact data quantity, but it is difficult to deal with the reachable critical points of the two clusters; the Gaussian clustering model is extremely sensitive to the initial value, which directly affects the performances of the algorithm, and the attribute features of the data itself can directly lead to the algorithm falling into local convergence in advance. These algorithms have great limitations for the analysis of learning behaviors, especially for the calculation and construction of learning interaction activities. The traditional clustering algorithms can not fully analyze and compare the data and relationships.

Aiming at the traditional clustering algorithms, this paper proposes an improved BIRCH clustering algorithm based on random walk strategy, that is, adding random walk to the data critical path clue process in clustering process. Through random walk, each learning interaction activity is modeled as a vertex, and the correlation of vertexes is mapped to the weight of the edge. The vertex and the edge form a graph, from which the related activity subset of the critical path of learning behaviors is found. The traditional BIRCH algorithm processes the hierarchical clustering of data, and the clustering feature tree is the typical data structure. The improved algorithm of BIRCH designed in this study is applied to the clustering analysis of learning interaction activities. Before the specific implementation of clustering, the key path of learning interaction activities is cued through random walk, and the data boundary of clustering analysis is reduced and determined, then BIRCH algorithm is implemented to accurately discover the individuality and group identity of learning interaction activities The main innovation of this processing method lies in the two-step processing scheme of data clustering. Combined with the requirements of path clues of learning interaction activities, random walk can improve the efficiency and accuracy of data analysis, reduce the interference of outliers, make up for the shortcomings of traditional BIRCH clustering, and improve the flexibility and effectiveness of the algorithm in dealing with large data sets.

## 2. Related Work

Clustering analysis plays an important role in the theoretical research and application of big data, which is widely used in business data, medical data, biological data, etc. Through clustering, abnormal data can be found, the standardization process of data to be detected can be promoted, the quality of data analysis can be improved. The hidden mechanisms and relationships in data can be obtained through data mining, serving for data prediction and feedback. After combing the relevant research results of data clustering, the following three series of cluster analysis are summarized:

(1) Optimization and improvement based on traditional clustering algorithm

In this aspect, the usual method is to optimize and improve a certain clustering algorithm (Martinez Martin P et al., 2020; Zhang Rong et al., 2020; Wan Jing et al., 2020), get a new clustering algorithm, and deduce the relevant target function and wave function needed in the clustering process. The algorithm is applied to cluster analysis of various symptom diagnosis, air quality, commodity, remote sensing, ionospheric radar wave and other data sets to verify the performance indicators of the improved algorithm, compare the quality of clustering results, and effectively identify the contribution of data features to different clustering units; or optimize the combination of traditional clustering algorithms (WAN Jing, etc, 2020), the algorithm prunes data and clusters data, and its function and performance are tested by individual or multiple data sets. The comprehensive application of various clustering algorithms can improve the defects of different algorithms to a certain extent. In the algorithm test, the accuracy rate, recall rate, F Score and other indicators are generally used.

(2) The fusion of random walk and traditional clustering algorithm

Random walk is applied to the design of clustering algorithms, and its feasibility and usability have been proved by some achievements (Zheng Wei et al., 2010; Wu Qiong et al., 2010; Li KC, & Wong BTM, 2020; Cormack S H et al., 2020). In this aspect, it is generally assumed that there is a random relationship in data, and the data is defined as a vertex in one graph. According to the similarity of the data, the transfer matrix of the random walk process is constructed. The random walk searches the relationships among vertexes, it gradually converges, the transfer matrix records the distribution of the vertexes to achieve the distribution clustering of the data. This fusion strategy, generally combined with clustering algorithms such as K-means, density probability model, particle swarm and so on. If the data comes from the network communities, it is suitable to use random walk to evaluate the hierarchies of the network community. Different hierarchy correspond to different diffusion time. In this case, we need to introduce a hierarchical clustering process.

(3) Clustering Research on education big data

Cluster analysis is applied to the research of online educational data to analyze the interaction process and activity distribution, and problems are found through the data (Sun Hongtao et al., 2016; Koh Y Y J et al., 2020; Yarygina O, 2020). The relevant data is generally taken from online platforms, and the research processes are divided into three aspects. First, data analysis is carried out according to the participation of learning interaction, clustering the data of teachers and learners, so as to judge the activities of learning interaction, such as the post, follow-up and comment of learning forum, or the participation of clustering test, clustering result and empirical analysis of correlation. The data clustering in this aspect is usually based on the clustering of the same interactive activity of multiple courses; the second is clustering analysis of multiple learning interactive activities of a specific course to obtain and test the effective learning behaviors, for example, by analyzing the online videos, online documents, reference materials, forum participation and other aspects, the learners are classified and divided, and the clustering list is calculated; Third, from the perspective of technologies, along the process of learning behavior formation. with the help of clustering algorithms, users mine the similarity model of learning behavior.

Considering the data analysis requirements of clustering technologies, , this study will select big data sets of learning behaviors based on these three series of research results, track learning behaviors with the random walk strategy, and then use BIRCH algorithm to realize hierarchical clustering of learning interaction process. Through the work of this study, we can provide research methods and experimental technology support for the follow-up similar research topics.

## 3. Problem Description of Learning Interaction Activities

Learning interaction activities are generated in the interaction process of the teachers, learners and courses. The activities include two forms: online and offline. Activities are used as a means to realize the learning processes and build learning behaviors. This study focuses on the clustering analysis and empirical study of learning interaction activities in online learning behavior big data set.

The big data set of learning behavior is taken from an online platform, and the complete data is extracted through the shared open interface of the platform. The data set scale reaches PB. The courses involve two categories: Social Science and STEM. Four and three courses are selected respectively, and the learning periods of the courses are four in total. Different courses, categories or periods of learning interaction activities are different. Learners' online learning processes form different learning interaction activities, which are stored as related description data. Category, content and period form the main clustering conditions for the study of learning interaction activities. Based on these conditions, the initialization and standardized classification before data clustering are completed. Figure 1 is the mapping relationships of these three clustering conditions.

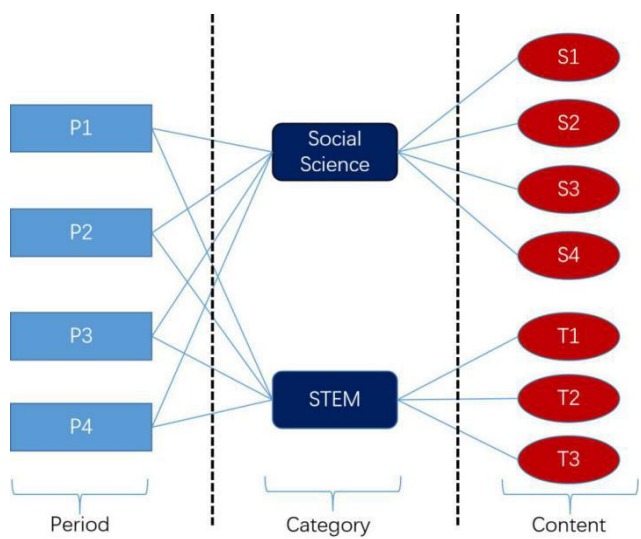

Figure 1 the Clustering Conditions of Learning Interaction Activities

The big data set of learning behaviors to be tested contains various learning interaction activities. Through statistical analysis of the data set, 20 kinds of learning interaction activities are obtained, as shown in Table 1. There are 107551 learners involved, which are distributed in different learning processes. Learners use appropriate learning interaction activities according to courses to form learning behaviors, a total of 16655280 records are generated.

The attributes of learning interaction activities include activity type, time, times and others, combined with course classification, content and period, random walk is used to track the learning interaction activities, find out the synchronous changes, build the key learning interaction path, and then carry out clustering analysis. According to different clustering conditions, the corresponding results are obtained through clustering analysis, and the rules and existing problems of learning behaviors are found from the results.

Table 1 Learning Interaction Activities

| Activity | Description |
| --- | --- |
| dataplus | Learn about real-time learning by dada analysis |
| dualpane | Parallel interaction of multi tasks |
| externalquiz | External quiz |
| folder | Classification of resources by folders |
| forumng | Learning interaction in the form of Forum |
| glossary | Online glossary for online reference |
| homepage | Visit personal homepage |
| htmlactivity | Learning interaction with hypertext |
| collaborate | Work together to achieve learning goals |
| content | Refer to online learning content |

| illuminate | Explain the questions provided |
| --- | --- |
| wiki | Visit wikipedia |
| page | Visit the pages involved in the learning process |
| questionnaire | Participate in questionnaire service |
| quiz | Quizzes in class |
| repeatactivity | Activities requiring multiple participation |
| resource | Resources for learners to download |
| sharedsubpage | Visit the shared subpage |
| subpage | Visit subpages through anchors |
| url | Visit uniform resource locator |

This study takes the data generated by learning interaction activities as the target, based on the previous research results, we classify and analyze learning interaction activities under the attributes, and demonstrates the key path and clustering trend of learning interaction activities in multiple dimensions. The execution sequence of the whole research work is shown in Figure 2, and the execution of each step will affect the execution of any subsequent step. The execution sequence of each step is as follows:

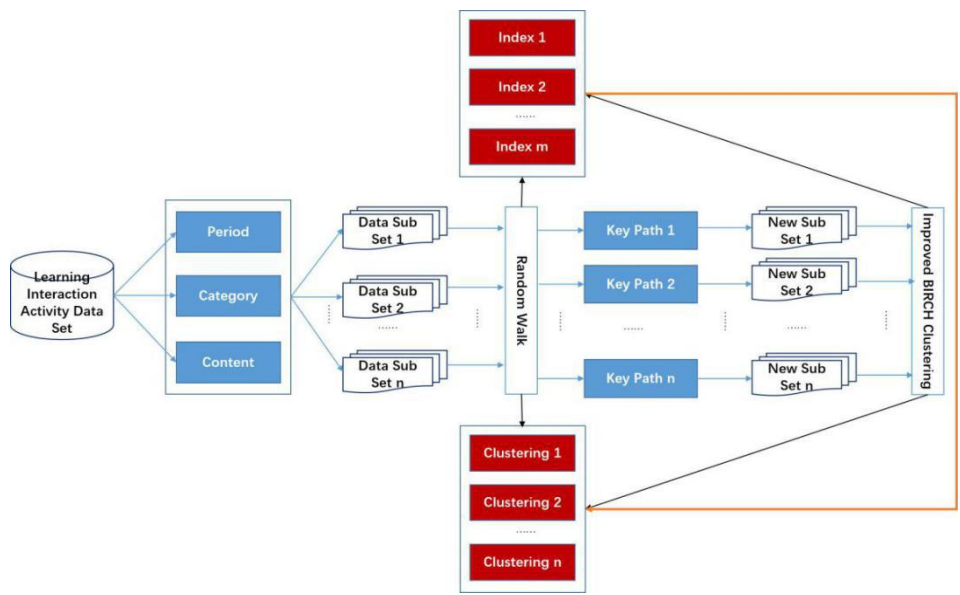

Figure 2 Clustering of Learning Interaction Activities

**Step 1:** The whole learning behavior big data set is divided into several relatively independent subsets according to the retrieval conditions (learning period, course category, course content, etc.);

**Step 2:** Random walk clue learning and tracking are carried out for each data subset, each data subset corresponds to a walk clue, forms the key path of learning interaction activities;

**Step 3:** The corresponding learning interaction activities are found from the data subset and form a new subset with the key path as the retrieval condition;

**Step 4:** the improved birch clustering algorithm is used to cluster each new subset. In order to verify the data analysis effect and clustering quality of this algorithm, the performance indexes of the improved algorithm and the traditional algorithm will be compared;

**Step 5:** According to the critical path of learning interaction activities, obtains the rules and features are obtained.

## 4. Improved BIRCH Based on Random Walk

In order to apply random walk and birch algorithm to clustering analysis of learning interaction activities, we need to consider their advantages and combine the features of data sets to optimize and improve the adaptability.

*4.1 Random Walk Model*
**Definition 1** Random Walk

Conserved quantities of irregular walkers themselves correspond to a diffusion law of transmission (Gosain A, & Sachdeva K, 2020). It is an ideal data model of Brownian motion, It can not deduce the potential execution steps and topological relationships according to the existing data. The path generated by the data is random and autonomous. Random walk is a data mining and tracking collection scheme in this case, which is usually applied to random process analysis.

The path of learning interaction activities forms a topological structure of graph, each activity corresponds to a vertex, and the relationship between two activities forms an edge. The topological structure of learning interaction activities can also be transformed into graph structure.

Supposing the graph $G$, the starting vertex is defined $V_0$. The walkers start from $V_0$, randomly select other $V_i$ as the walking target, then select $V_i$ as the starting vertex after arriving, reselect another vertex $V_j$ as the walking target ($i \neq j$), and so on. The set of vertices that walkers pass through constitutes the random walk route $G'$ ($G' \subseteq G$) of graph topology,.

Before designing a random walk algorithm for learning interactive activities, the definition, rules and algorithm of random walk are given.
**Definition 2** Random Process

Supposing $S_k(k=1,2,...n)$ is a group of random tests. Each test produces a waveform curve (curve is sample function) with time as an independent variable. The curve is recorded as $x_i(t)$, $t$ represents time, then the population $\{x_1(t), x_2(t),..., x_n(t),...\}$ of all possible results constitutes a random process $\xi(t)$, the population composed of all sample functions is called a random process.

$\xi(t)$ has two features, one is a population of whole samples, every samples is a wave function of $t$, which describes the occurrence of samples with probability; The other is that $\xi(t_k)$ observed at $t_k$ is a random variable without $t$. The random process can be regarded as a family of random variables depending on time parameters. The random process is determined by random variables and time functions.
**Definition 3** Markov Chain

Markov chain is a set of discrete random variables with Markov property, and a set of random variables $X = \{X_n : n > 0\}$ with one-dimensional countable set $s$ in probability space $(\Omega, F, P)$. If the value of random variable belongs to countable set, $X = s_i, s_i \in s$, when the conditional probability of random variable satisfy $p(X_{t+1} | X_t, ..., X_1) = p(X_{t+1} | X_t)$, $X$ is Markov chain. Countable set $s \in Z$ is called state space, and the value of Markov chain in state space is state.

**RW Algorithm**

---

**Initialization**: Let $f(x)$ be a multivariate function (sample) with multiple learning interaction activities, $X = (x_1, x_2, ..., x_n)$ is a dimension vector.

**Step1:** Given the initial iteration point is $X$, the initial transfer step length $\lambda$ and the control accuracy $\varepsilon$ ($\varepsilon$ is a very small positive number as the control threshold);

**Step2:** Given the number of iterations $N$, $k$ is the current number of iterations, and the initial value is 1;

**Step3:** If $k < N$, a $n$ dimension vector randomly $u = (u_1, u_2, \cdots, u_n), (-1 < u_i < 1, 1 \leq i \leq n)$ in $(-1, 1)$ is generated. After standardization, it is 

$u' = \dfrac{u}{\sqrt{\sum_{i=1}^{n} u_i^2}}$. Let $x_1 = x + \lambda u'$, the first step of random walk is completed;

**Step4:** if $f(x_1) < f(x)$, the first vertex better than the initial value is found. Let $k = 1$, $x_1 = x$, return to Step2; if not $f(x_1) < f(x)$, return to Step3;

**Step5:** If no better vertex is found in successive $N$ times, the key path of learning interaction activities is in the $N$-dimension space with the current optimal solution as the center and the current step length as the radius (either plane or sphere). If $\lambda < \varepsilon$, RW algorithm will end; if not $\lambda < \varepsilon$, it will return to step1 and start a new round of random walk.

---

Transition probability is an important part of Markov chain. If Markov chain is composed of $m$ States, starting from any state and passing through any transition, one of the $m$ States will inevitably appear, the transition between States is called transition probability. $\{X_n, n \geq 0\}$ is set as discrete-time Markov chain, for any $m \geq 0, n \geq 1, i, j \in E$, $E$ is the state space, $P_{i,j}(m, m+n) = P\{X_{m+n} = j | X_m = i\}$ is called step transition probability from state $i$ to $j$ after $n$ steps, which is called $n$ step transition probability, one step transition probability is $P_{ij}(m, m+n) = P\{X_{m+n} = j | X_m = i\}$.

On the basis of Definition 1, 2 and 3, we use random walk to trace learning interaction activities and and extract the corresponding activities of critical path from the perspective of global optimization. The algorithm can ensure that the data processing is not easy to fall into premature convergence. The algorithm flow is described as RW algorithm.

*4.2 Improved BIRCH Clustering Algorithm*

BIRCH(Balanced Iterative Reducing and Clustering Using Hierarchies) is a balanced iterative planning and clustering algorithm realized by hierarchical method. The implementation goal of the algorithm is to complete clustering only by scanning the data set one time with high efficiency and accuracy(Marichamy V S , & Natarajan V, 2020).

The data structure of BIRCH is similar to the balanced B + tree, that is, the clustering feature tree( $CF$ Tree), each node in the tree is composed of several clustering features, which are divided layer by layer until the leaf nodes. The leaf nodes also have several $CF$. $CF$ of non leaf nodes have pointers to the next layer of hierarchical nodes, All leaf nodes are linked through the data structure of the double linked list, as shown in Figure 3.

**Definition 4** $CF$ : Each $CF$ is represented as a triple, $CF = (N, LS, SS)$. $N$ represents the number of samples, $LS$ is the sum vector of each feature dimension, $SS$ is the square sum of each feature dimension. $CF$ has a linear relationship and the relevant parameters can be superposed, that is $CF_1 + CF_2 = (N_1 + N_2, LS_1 + LS_2, SS_1 + SS_2)$.

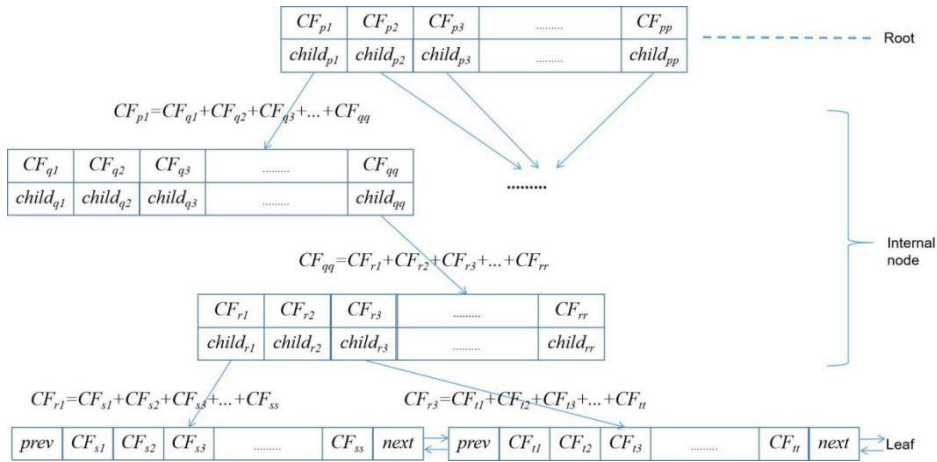

Figure 3 Structure of Clustering Feature Tree

Based on the definition of clustering features and the features of feature tree, it is the core idea of BIRCH to establish $CF$ tree for all the data set samples of learning interaction activities. The output result after BIRCH clustering is a number of $CF$ nodes, the samples of each node are clusters, and BIRCH clustering of learning interaction activities is the process of building a $CF$ tree. The main process is shown in the BIRCH clustering algorithm.

During the design of the birch clustering algorithm, some exceptions need to be considered, mainly including:

(1) The course content and period are the two index conditions to differentiate the sample set

of learning interaction activities, which can realize the parallel clustering analysis;

(2) At the beginning of building $CF$ trees of learning interaction activities, we need to filter and remove some abnormal $CF$ nodes. Generally, these nodes are very few sample points. For the tuples which are very close to $CF$, we should perform the merging;

(3) First, we use random walk to retrieve the key path of learning interaction activities and preprocess it with clues, and then cluster all $CF$ tuples, so that we can get a better $CF$ tree, so as to eliminate the unreasonable tree structure caused by the large scale of samples, and avoid the serious impact of the number limit on the $CF$ tree structure, resulting in improper structure decomposition.

(4) Through the algorithm to generate the center points of all $CF$ nodes, all the sample points are classified and clustered according to the absolute value of the distance between the center point and the threshold, further reducing the unreasonable clustering results caused by some restrictions during $CF$ tree construction.

Before BIRCH clustering, the key path of the data set is threaded through random walk to improve the matching of the appropriate samples and the clustering speed, and the nodes of $CF$ tree are added, deleted and merged very quickly, which also enables the initial data set to be effectively screened and extracted in advance, identify abnormal data and interference data as early as possible, and improve the efficiency and quality of data analysis.

**BIRCH Improved Algorithm**

---

**Initialization :** According to different analysis requirements, the data of learning interaction activities are divided into different data subsets.

**Step1:** RW algorithm is called to realize the key path of the data subset. The key path is mapped to the secondary screening conditions of learning interaction activities, and the screening result is the data set to be clustered;

**Step2:** The first sample point is read from the data set and filled into a new $CF$ triple as the root node;

**Step3:** If the radius of the sphere for $CF$ as the central base point is less than the threshold $T$ after the new sample is added, all $CF$ triples on the path will be updated, otherwise, turn to Step4;

**Step4:** If the number of $CF$ of the current leaf node is less than the threshold $L$, a new $CF$ will be created and put in a new sample, then the new $CF$ is taken as the leaf node, and all $CF$ triples on the path are updated, otherwise go to Step5;

**Step5:** Split the current leaf node into two new leaf nodes, and select all the $CF$ triples in the original leaf node corresponding to the two $CF$ triples with the farthest distance from the sphere as the first $CF$ node of the new leaf node;

**Step6:** According to the principle of the absolute value of the difference between the threshold value and other triples, new samples are returned to the corresponding leaf nodes. Iteratively, the internal nodes are checked whether or not need to be split, and if so, iterate the split process of Step4 and Step5.

---

## 5. Experiment

### 5.1 Configuration

The environment of this research experiment is: Intel (R) core (TM) i7-8550 CPU@1.80GHz

1.99ghz, 16g memory, 64 bit operating system is Win10, x64 based processor. The executable code of the algorithm is written with python3.7.

The whole data analysis process is divided into two steps: one is to search and clue the key path of the learning interaction data set under different conditions through RW algorithm, and then filter the learning interaction data to form multiple data sets to be clustered under the condition of clue results; the other is to use the improved BIRCH algorithm to perform the clustering process on different data sets.

*5.2 Random walk experiment*

Before the code design of random walk, the whole data set can be directly divided into two data subsets under the condition of course category. On this basis, the key path of random walk is cued from different course contents and different learning periods. The course content may involve four periods, and the execution route of random walk should be restricted by course content and period. RW algorithm executable program acts on two data subsets respectively, the time complexity is $O(n \log n)$. The topology of Figure 4 is obtained during the operation of the program.

In Figure 4, the first period forms the key paths of two subgraphs. According to the statistical analysis of online learning platform and learners' learning behaviors, only one of the four courses of Social Science is involved in the course selection, and only one key path of learning interaction activities is extracted from the generated data, which is consistent with the actual situation; The key path of two courses is extracted from the program analysis of STEM. In the same way, the second, third and fourth learning period are implemented accordingly.

Compared with the eight subgraphs in Figure 4, Social Science courses are the most intensive in the fourth period. In addition to the first period, the other periods of STEM courses have learning interaction records. From the scale of data set, the STEM interaction data is much larger than that of Social Science, and the online opening and learning of STEM are more sufficient, and the interaction process is more frequent.

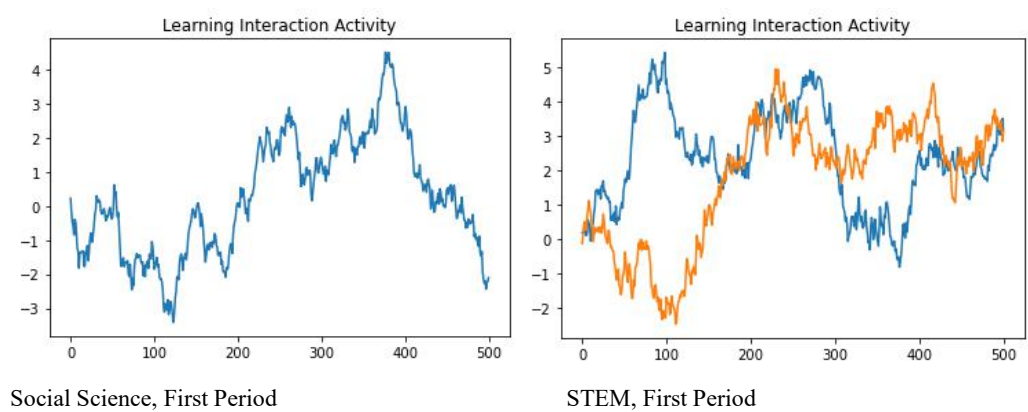

Social Science, First Period　　　　　　　　　　　　STEM, First Period

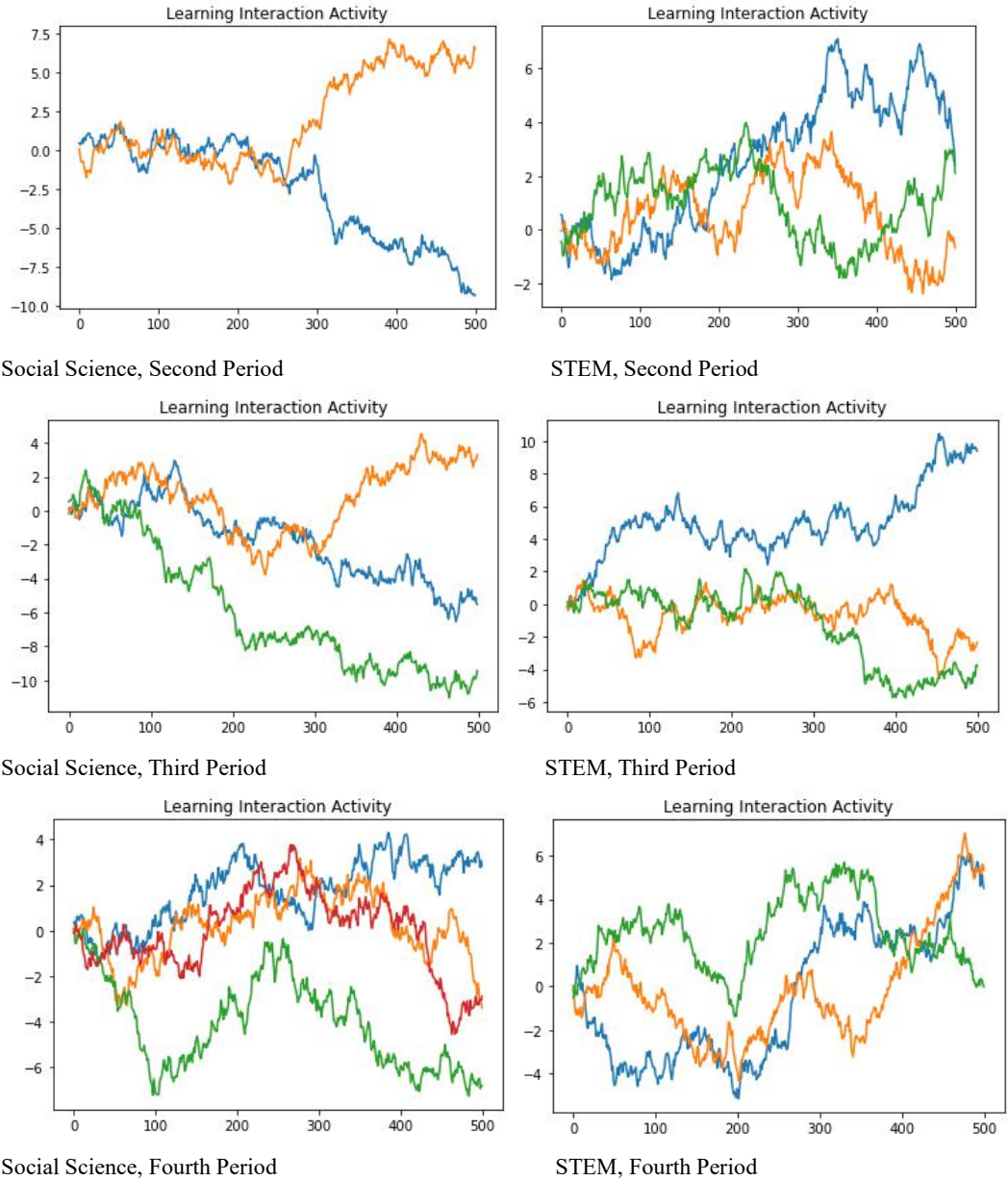

Social Science, Second Period    STEM, Second Period

Social Science, Third Period    STEM, Third Period

Social Science, Fourth Period    STEM, Fourth Period

Figure 4 random walk of key paths of seven courses in four periods

Taking the random walk of 8 key paths in Figure 4 as the data retrieval condition, the sample points corresponding to seven courses in four periods are retrieved and extracted from two data subsets respectively. The data obtained is a file structure that retains the original data set structure and type, and is saved in CSV format. The time complexity of the corresponding python program is $O(n)$, the double limitation of course and period, the two data subsets are mapped to 22 CSV files, and their scales are shown in Table 2. In Table 2, the data of learning interaction activities generated by STEM courses is larger, and the participation of the same course in different periods is more balanced. Only S2 in Social Science has online learning behaviors in four periods, and the other three courses only occupy two periods.

Table 2 Original data Scale of Different Courses in Different Periods

| Category | Course | First Period | Second Period | Third Period | Fourth Period |
| --- | --- | --- | --- | --- | --- |
| Social Science | S1 |  | 180,982 |  | 169,316 |
| | S2 | 403,266 | 452,638 | 273,236 | 438,424 |
| | S3 | | | 496,181 | 711,646 |
| | S4 | | 145,995 | 123,319 | 117,859 |
| STEM | T1 | 536,837 | 680,806 | 379,942 | 568,901 |
| | T2 | | 356,262 | 202,224 | 402,947 |
| | T3 | 946,765 | 1,172,101 | 685,274 | 1,210,359 |

*5.3 Clustering Experiment of Improved Birch Algorithm*

To realize the BIRCH clustering of learning interaction activities, the data should be formed first. On the basis of Table 2, data transposition and normalization are realized for subsets with learners as record items and learning interaction activities as data feature structures. According to the structural features of the learning behavior data set, click rate is regarded as the statistical index, and the double condition retrieval and perspective calculation of click rate are realized with learners and learning interaction activities as the common limitations. The execution program of the whole analysis process is realized by python3.7 and the visual result is formed. The time complexity is $O(n^2)$. The feature number and scale of the subset of learning interaction activities are shown in Table 3.

Table 3 Feature Number and Scale of Learning Interaction Activities

| Category | Course | Feature Number \| Scale | | | |
|---|---|---|---|---|---|
| | | First Period | Second Period | Third Period | Fourth Period |
| Social Science | S1 | | 4 \| 378 | | 4 \| 357 |
| | S2 | 10\|1527 | 10 \| 1870 | 10 \| 1294 | 10 \| 1921 |
| | S3 | | | 9 \| 1681 | 9 \| 2302 |
| | S4 | | 7 \| 895 | 7 \| 773 | 7 \| 698 |
| STEM | T1 | 11 \| 1214 | 10 \| 1768 | 10 \| 1116 | 10 \| 1647 |
| | T2 | | 11 \| 964 | 11 \| 624 | 11 \| 1097 |
| | T3 | 14 \| 1510 | 16 \| 2098 | 15 \| 1563 | 16 \| 2121 |

In Birch clustering, the variance of cluster is set to 0.3 ~ 0.4, and the cluster centroid coordinates are distributed on the function y =x. The program respectively acts on 22 learning interactive data subsets in Table 3, and the clustering results of Figure 5 are obtained. The location of each cluster distribution sub graph is respectively mapping to the data set listed in Table 3, the lower right corner of each distribution sub graph corresponds to the number of clusters.

In Figure 5, each sub tree whose node is the root node corresponds to a cluster. For cluster analysis, take S1 and T3 courses as examples. From the clustering results of the third and fourth periods of S1, it can be seen that the learning interaction activities of S1 are the clustering of four features, and each feature is the root node of the subtree obtained from the split of sample points. Due to the small scale of the data set, the distribution of sample points of each subtree is discrete. No matter in the second or in the fourth period there are four clusters; In the fourth period, T3 has 19 kinds of learning interaction activities. After clustering only five clusters are generated, that shows that the learning interaction activities are relatively concentrated. There are differences in learning interaction activities of the same course in different learning periods, and there are differences in the number of clusters after clustering, which shows that learners' participation in learning interaction activities has a tendency, but learning content does not play a decisive role.

The clustering results in Figure 5 also illustrate an important conclusion, that is, the learning behavior group tendency of STEM is obviously stronger than that of Social Science, there are many kinds of learning interaction activities in STEM courses, and the amount of features is large, but the learners' preferences for participating in learning interaction activities are similar. For the online learning mode, the learning interaction activities of STEM courses are more diverse, and the participation is more frequent, which is conducive to the formation of significant learning group features. The learning process and learning behavior are associated and transitive for individuals.

By using the improved BIRCH algorithm clustering analysis, the $CF$ splitting is tracked, and the root nodes of subtree when each data subset sample points are split are obtained. Table 4 is results after calling the statistical program, and the sequence of learning interactive activities is defined as the sequence of root nodes. The data of each cell in Table 4 corresponds to the clustering result graph of the same location in Figure 5. The clusters are formed by the combination of these features or the cascade of features.

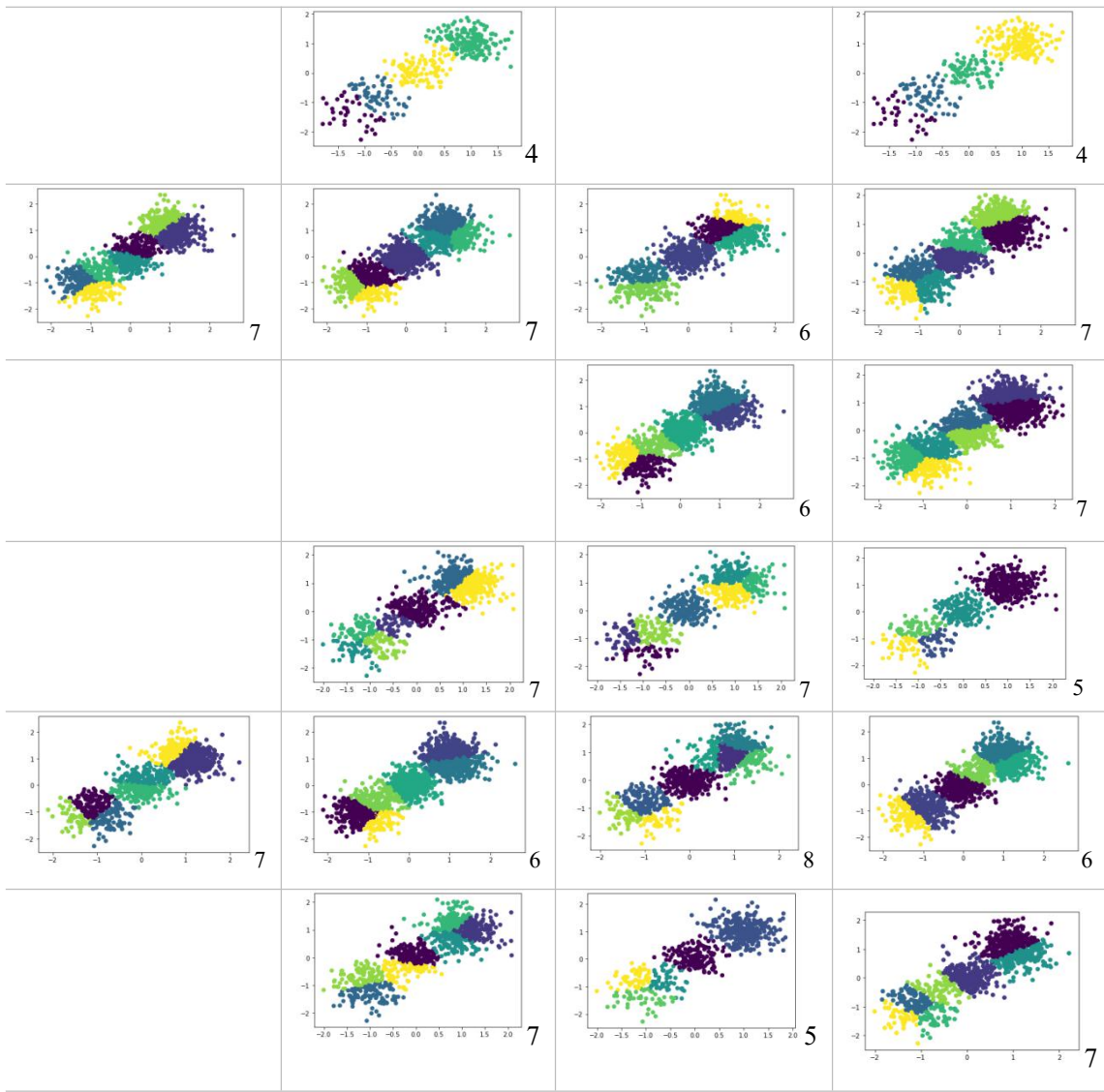

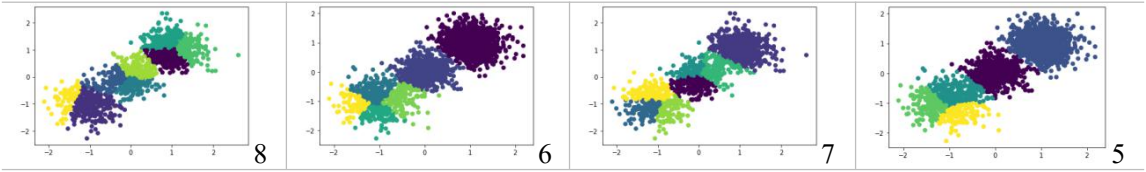

Figure 5 BIRCH Clustering Results of Learning Interaction Activities of Different Courses in different learning Periods

Table 4 Learning Interaction Activities corresponding to Root Nodes of $CF$ Trees

|  |  |  |  |
|---|---|---|---|
|  | homepage\|forumng content |  | subpage\|content resource |
| subpage\|resource homepage\|url forumng\|quiz | subpage\|resource homepage\|quiz forumng\|url\|content | homepage\|forumng glossary\|subpage resource\|url | homepage\|forumng subpage\|quiz contentquestionnaire |
|  |  | homepage\|forumng subpage\|url quiz\|resource | homepage\|forumng subpage\|url page\|quiz |
|  | homepage\|forumng content\|subpage quiz\|resource | homepage\|forumng content\|quiz subpage\|resource | content\|homepage quiz\|forumng subpage |
| subpage\|resource homepage\|forumngurl | homepage\|subpage forumng\|externalquiz wiki\|resource | homepage\|forumng wiki\|url\|subpage resource\|externalquiz | homepge\|forumng subpage\|collaborate url\|externalquiz |
|  | url\|homepage forumng\|content\|wiki | url\|homepage forumng\|wiki\|content | homepage\|forumng url\|wiki\|content |
| subpage\|content\|urlhome page\|forumng page\|questionnaire dataplus | subpage\|content\|url homepage\|forumng questionnaire dataplus | subpage\|content homepage\|forumng url\|questionnaire dataplus | subpage\|content homepage\|forumng url\|questionnaire dataplus |

From Table 4, it can be seen that each course has different learning interaction activities in different learning periods, and the order in which different levels and characteristics are divided into root nodes is also different. However, as the root node, the learning interaction activities are similar to the same courses. After statistical analysis of data, the correlation between learning interaction activities also has the same trend. Take S2 and T3 as examples, we calculate the correlation between homepage and forumng in four learning periods, and get Table 5. The mark S2-1 in the table represents the learning interaction activity data in the first period of S2 course, and the meaning of other marks is similar. These two courses are positively and strongly correlated in the four learning periods. In the same way, we can analyze the correlation between other learning interaction activities, and the experiences of learning interaction activities as the key path have strong positive correlation.

Table 5 Example of Correlation of learning interaction activities of $CF$ Tree Root Node

| Correlation | S2-1 | S2-2 | S2-3 | S2-4 | T3-1 | T3-2 | T3-3 | T3-4 |
|---|---|---|---|---|---|---|---|---|
| homepage ↔forumng | 0.8314 | 0.8487 | 0.8591 | 0.7894 | 0.7978 | 0.8072 | 0.8243 | 0.8492 |

*5.4 Experimental Conclusion*

the experiments of random walk and improve BIRCH have analyzed and tested the whole learning behavior big data set, we can see that the fusion application of this algorithms is effective, that is, the key path of learning interaction process is formed after running random walk, which can serve the root node decision-making selection and adaptive transfer of clustering feature subtree, and the clustering processes provide appropriate node splitting strategy. This conclusion is of great significance to the study of learning interaction activities. Random walk plays a leading role in the clustering process of improved BIRCH. This processing method can greatly improve the data analysis quality and efficiency of BIRCH.

## 6. Algorithm Performance Comparison

In order to verify the clustering performance of the improved BIRCH based on random walk, 22 data subsets in Table 3 are clustered using the traditional BIRCH algorithm, and the two algorithms are compared from five performance indicators, namely *Precise*, *Accuracy*, *Recall*, *F Score* and *algorithm time consumption*. The calculation model of the first four indicators is related to four statistics: $FP$ represents the number of samples that are actually negative but predicted to be positive, $TN$ represents the number of samples that are actually negative and also predicted to be negative, $TP$ represents the number of samples that are actually positive but predicted to be positive, $FN$ represents the number of samples that are actually positive but predicted to be negative (Marichamy V S, & Natarajan V, 2020).

*Precise* is the probability of the actual positive samples in all the predicted positive samples, the larger the value, the better. The calculation model is expressed as Formula 1:：

$$Precise = \frac{TP}{TP + FP} \quad \text{(Formula 1)}$$

*Accuracy* is the probability of all correctly predicted samples in the total sample. The larger the value, the better. The calculation model is expressed as Formula 2:

$$Accuracy = \frac{TP + TN}{TP + TN + FP + FN} \quad \text{(Formula 2)}$$

*Recall* refers to the probability that the actual positive sample is predicted to be a positive sample. The larger the value, the better. The calculation model is expressed as Formula 3:

$$Recall = \frac{TP}{TP + FN} \quad \text{(Formula 3)}$$

*F Score* has both *Accuracy* and *Recall*. The goal is to achieve the maximum at the same time, so as to achieve a balance. The calculation model is expressed as Formula 4:

$$F\ Score = \frac{2 \times Precise \times Recall}{Precise + Recall} \quad \text{(Formula 4)}$$

In the comparative experiment, 22 data subsets were divided into two categories: Social Sciences and STEM. Each category included 11 data subsets. For each index, the process tracking and result comparison of data analysis are carried out. The result of index value is shown in Figure 6 and Figure 7, each figure includes four subgraphs of index. As can be seen from the line of the distribution of the four indexes of *Precise*, *Accuracy*, *Recall* and *F Score* in Figure 7 and figure 8, the algorithm in this paper (BIRCH Improved) has advantages, especially for the larger scale and the more the number of features of the data subset, he improved algorithm is superior. The interval comparison lines of the two algorithms in Figure 7 is more obviously than that of Figure 6. The

improved BIRCH based on random walk is more conducive to the clustering analysis of large-scale data sets

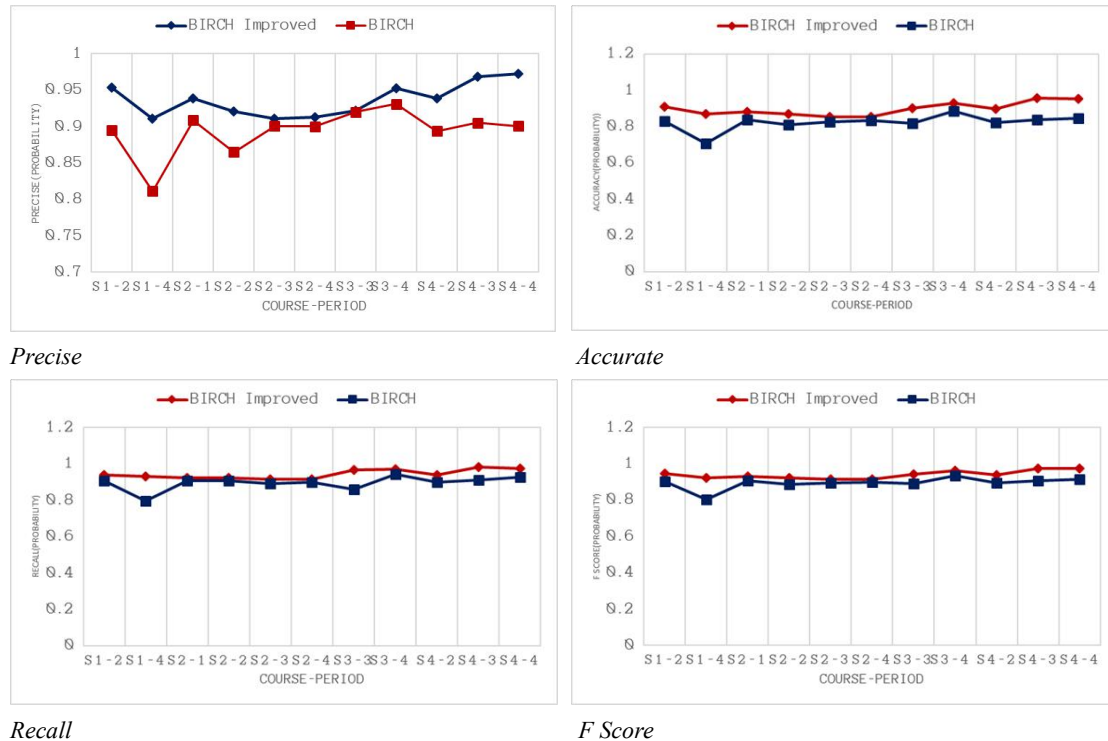

*Precise*  *Accurate*

*Recall*  *F Score*

Figure 6 Social Science: the Comparison of BIRCH Improved and BIRCH

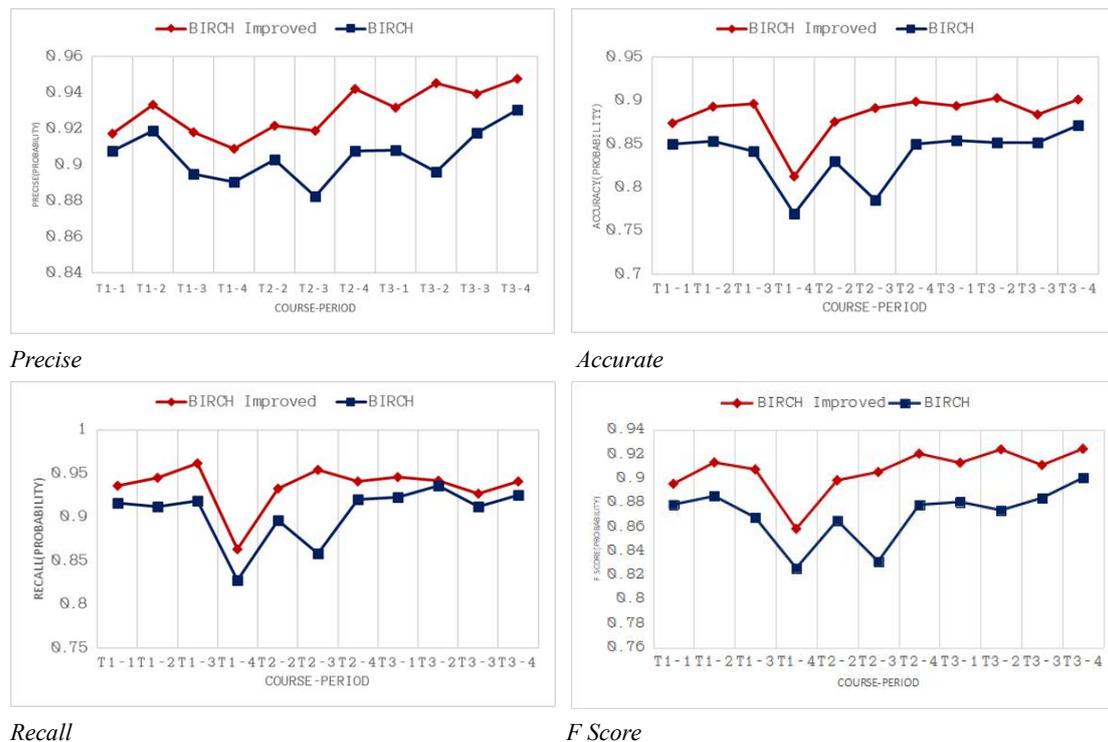

*Precise*  *Accurate*

*Recall*  *F Score*

Figure 7 STEM: the Comparison of BIRCH Improved and BIRCH

　　The clustering time consumption of 22 data sets by BIRCH Improved and BIRCH is shown in Table 6. For different data sets to be detected, the time consumption of BIRCH Improved is less than BIRCH in different degrees. When there are more features of data, the time-consuming gap is larger. The BIRCH Improved is also conducive to large-scale and multi feature data sets in the efficiency of program operation. The random walk strategy improves the initialization and standardization quality of data sets, strengthens the positioning accuracy and suitability of targeted data samples, and accelerates the clustering process (Shi Lingjuan, & Huang Decai, 2020; Liu Jiayao, & Wang Jiabin, 2020; fan Zhongxin, etc., 2019; Izadi H et al., 2020).

Table 6 Time Consumption Comparison of BIRCH Improved and BIRCH

| COURSE -PERIOD | Running Time (/s) | | COURSE -PERIOD | Running Time (/s) | |
| --- | --- | --- | --- | --- | --- |
| | BIRCH Improved | BIRCH | | BIRCH Improved | BIRCH |
| S1-2 | 2.10 | 2.66 | T1-1 | 9.07 | 10.22 |
| S1-4 | 2.00 | 2.14 | T1-2 | 8.81 | 9.48 |
| S2-1 | 7.15 | 8.04 | T1-3 | 7.96 | 8.85 |
| S2-2 | 7.27 | 8.55 | T1-4 | 8.59 | 9.99 |
| S2-3 | 6.09 | 6.63 | T2-2 | 8.82 | 9.83 |
| S2-4 | 7.44 | 7.98 | T2-3 | 8.11 | 9.79 |
| S3-3 | 7.17 | 7.66 | T2-4 | 9.33 | 10.44 |
| S3-4 | 8.84 | 9.08 | T3-1 | 10.09 | 12.38 |
| S4-2 | 3.22 | 4.85 | T3-2 | 13.39 | 16.61 |
| S4-3 | 3.01 | 4.10 | T3-3 | 12.65 | 14.93 |
| S4-4 | 2.83 | 3.53 | T3-4 | 17.88 | 20.95 |

## 7. Research Summary and Decision Suggestions

In this study, cluster analysis is carried out for the group tendency of online learning behaviors, and the learning behaviors are described systematically and related through multi-dimensional learning interaction activities. Firstly, we select the learning behavior big data set with multi periods and multi courses, and the algorithm technology is used to initialize and standardize the data set. According to the data classification conditions, the data set is properly differentiated or integrated, and 22 data subsets that can be used for clustering analysis are obtained. Secondly, based on the topological correlation and dependence of different learning interaction activities, we design the improved algorithm based on random walk to achieve the topology of retrieval route and data clustering results of key learning interaction activities. Furthermore, the comparative index experiments are carried out for BIRCH Improved and BIRCH. The experimental results show that the improved birch algorithm in this paper has advantages in Precise, Accuracy, Recall, F Score and program time consumption. The index values of the analysis are all within the obvious advantage range. Our algorithm can effectively guarantee the feasibility and reliability of the data clustering process and results, and can be used as the basis for the research of big data-driven education and teaching decision-making.

This section makes a conclusion summary of the whole clustering research process of learning behaviors, and provides appropriate decision-making suggestions for the problems and conditions caused by data analysis.

(1) Course category affects online learning behaviors.

In this study, the courses are divided into two categories: Social Sciences and STEM. The amount of data generated by online learning behaviors of these two categories is significantly different. The learning behavior data of STEM courses are generally higher than that of Social Sciences courses, and the corresponding learning interaction activities of STEM courses are also more diverse. The types of learning interaction activities of social science courses are relatively single and centralized, the clustering results of learning behavior data are relatively scattered and sparse, the tendency of learners is not obvious, and learning interaction activities are single, but they form more clustering clusters, which shows that online learning means of Social Science courses learners are more spontaneous, and the results are extracted from key learning interaction activities. So Social Science learners generally tend to take part in online materials, glossaries, terms, content and other activities, and rarely participate in collaborative and interactive activities. Online learning mode is more suitable for learners as an auxiliary means of curriculum learning (Lucy González, & Lugo C, 2020). Compared with Social Science, online learning of STEM is another scenario. The key learning interaction activities are mainly reflected in online interaction, discussion and collaboration. Online participation is intensive, and online learning is the more mainstream learning mode for STEM learners.

Suggestions and Decisions of education and teaching: the course category affects online learning behaviors, which determines that online learning cannot be "one size fits all" (Wu Fati et al., 2019; Wang Xuenan, 2019; Chen Dexin et al., 2019). Online learning mode is not conducive to the whole process of education and teaching, and it is more suitable to be an appropriate supplement. From the perspective of data analysis, STEM courses are more conducive to the construction of online learning behaviors than Social Science courses. In the process of education and teaching, "online" learning can not replace "offline" teaching (Lucy González, & Lugo C, 2020). Therefore, in the process and means of education and teaching, we need to consider the category of courses and flexibly implement the online and offline mixed mode according to the actual situation of learners.

(2) Course content affects online learning behaviors.

The learning behaviors of different courses are similar to each other, but the differences are also obvious. For example, there are common learning interaction activities among the four courses of Social Science, but there are also typical differences. The same is true for the three courses of STEM. Even if they belong to the same course category and different course contents, they can produce their own learning behaviors. The scale of online learners is also different, and the degree of participation is also different. Among the seven courses, T3 course has the largest degree of participation and the most types of learning interaction activities, but the clusters generated by clustering are relatively small, and learners' learning behaviors have formed a relatively stable and centralized mode, that is, the course has obvious characteristics of group learning, which can play a faster guide for new learners to participate in the course learning, it is also conducive to establish a systematic self-study mode in a short time, and realize the effective development of online learning and active optimization of learning processes.

Suggestions and Decisions of education and teaching: better online learning group conformity plays a strong role in promoting the construction of course learning model, which has been proved many times in the research of similar learning behaviors (Viloria A, & Naveda A S, 2020; Miah S J et al., 2020; Xia Xiaona, etc, 2018), this study also reached the same conclusions through cluster analysis of learning interaction activities with high participation. Education and teaching need to teach students in accordance with their aptitudes, but learning methods and teaching methods need to build a benign group guidance mechanism, so that learners can form an independent tracking and appropriate adjustment of the self-active learning processes. Therefore, in the process and means of education and teaching, it is necessary to consider the contents of the courses, combine the existing experiences and lessons, start from the results of learning analysis, and gradually build a sustainable education and teaching feedback measures and mechanism, so as to guide learners to build their active awareness of self discover learning, build learning and adjust learning.

(3) Learning period affects online learning behaviors.

It can be seen from the clustering analysis results of learning behavior data of four periods of seven courses that the key learning interaction activities of the same course in different learning periods are not the same, and the differences are more obvious in different periods of the same course and other courses, and the number of learning interaction activities of the same category or the same course in different periods is also different, that is reflected in the data analysis of different periods of seven courses, that is, different periods, the same learning content, and the choice of learning interaction activities will change. On the one hand, it is related to the learners themselves; on the other hand, the changes of learning objectives, lecturers, teaching methods, teaching guidance cases and experiments corresponding to different learning periods are all the influencing factors of different learning behaviors (Jude Osakwe et al., 2020; Wunan Zhong, 2019; Meng Dongdong, etc., 2020). Learning behavior is flexible, subject to some factors, teaching according to aptitude, but also need to do teaching according to time, place and learning, etc.

Suggestions and Decisions of education and teaching: for the design of education and teaching mode, time factor cannot be ignored. Other factors related to time factor, such as learners, instructors, education and teaching objectives, learning objectives, etc., also need to be considered. The existing learning behaviors data can inspire the potential learning behaviors, especially the favorable learning and interaction methods, which can be directly inherited among learners and reduce the time difference for learners to integrate into learning. Therefore, the periods of education and teaching, first of all, needs to do a good job in the analysis and prediction of historical data, so as to achieve the early assessment of the learning process; second, we need to objectively assess the current situation of learners' attribute characteristics, and develop appropriate teaching methods and learning guidance strategies; third, while the learning process is advancing, needs to timely trace and analyze the existed data, then carry out the formulate learning behavior optimization plan.

(4) Activity association affects online learning behaviors.

Table 5 is an example of the conclusion that there is a strong correlation between key learning interaction activities. For S1 and T3 courses, homepage and forumng are closely involved. The participation of one of these two activities will lead to the participation of the other. In addition, through data analysis, the two learning interaction activities of questionnaire and dataplus involved in T3 are closely related each other. These two activities rarely appear in other courses, but in the fourth period of T3, these two activities almost appear at the same frequency. It can be seen that these two activities are necessary for T3. The correlation of learning interaction activities will lead to corresponding learning behaviors, which can be obtained through the analysis of the correlations between learning interaction activities.

Suggestions and Decisions of education and teaching: the correlations of learning interaction activities enable different activities to establish the opposite participation. Therefore, the methods, means and models adopted in the process of education and teaching will have an impact on learning behaviors and bring about possible changes in learning outcomes. Educators should track the learning processes through appropriate data analysis technology and means, guide the use of relevant functions of online learning platform through interaction, cooperation, teaching, question-answering, investigation and others in the teaching processes. From the perspective of platform construction, the presentation of functional views should follow the layout and deployment of relevance and proximity to provide a more friendly way of human-computer interaction.

(5) Algorithm design affects the analysis of learning behaviors.

There is no fully universal data analysis algorithm, nor fixed data structure and attribute characteristics. Many data analysis functions of software are limited, and there are many restrictions for the data itself. If the traditional algorithm is directly applied to the data set, it can get the inefficient data analysis process and low-quality results. The analysis work needs to start from the data itself, according to different demonstration objectives, optimize and design appropriate algorithms. Therefore, based on the complexity and particularity of learning behaviors, this study designed an improved BIRCH clustering algorithm based on random walk strategy. Through the measurement of *Precise*, *Accuracy*, *recall*, *F score* and *algorithm time consumption*, the performance comparison between BIRCH improved BIRCH was carried out to demonstrate the effectiveness of the improved algorithm, the feasibility and reliability of data clustering results.

Suggestions and Decisions of education and teaching: data analysis in the process of education and teaching should start from the data itself to determine methods and technologies. It is not advisable to prune data based on algorithms, which will cause different degrees of distortion and deviation, lose the significance of data analysis and data-driven decision-making, and ensure the integrity and reality of data is very important. According to the decision-making requirements, the effective data boundary can be located, and the complete data can be extracted by using technical tools and algorithm programs. Data analysis based on algorithm and model design should become an important implementation idea of data-driven decisions. That Data serves decisions is an important part of the development of education science and data science (Zhao Yixia et al., 2019; Bao haogang et al., 2019). Therefore, the corresponding data and appropriate algorithm technologies are required for education and teaching to achieve accurate and delicate decisions.

## 8. Outlook

It is the key research scheme and experimental strategy to infiltrate the statistical model and algorithm program into the research of education and teaching, which plays an important role in promoting the development of education science. The follow-up study and analysis work will focus on the needs of education and teaching decisions, deeply argue learning influence factors and process description characteristics with multi angle perspectives, so as to put forward relevant problem hypotheses, build factor models, design data modeling methods and algorithm programs, comprehensively use appropriate technical means, construct the analysis scheme of problems and get verification conclusions, which is of great significance and promotion value to theoretical research and practical application of data-driven education and teaching.

## 9. References

Christian Fischer, Zachary A Pardos, & Ryan Baker (2020). Mining Big Data in Education: Affordances and Challenges. *Review of Research in Education*. 44(1): 130-160.

Dahdouh, K. , Dakkak, A. , Oughdir, L. , & Ibriz, A. . (2020). Improving Online Education Using Big Data Technologies. *The Role of Technology in Education*.

Liu Min, & Zheng Mingyue. (2019). Learning Analytics and Learning Resources Personalized Recommendation in Smart Education. *China Educational Technology*(9), 38-47.

Chen Kaiquan, Zhang Chunxue, Wu Yueyue & Liu Lu(2019). Multi-modal Learning Analysis, Adaptive Feedback and Human-computer Coordination of Artificial Intelligence in Education（EAI）. *Journal Of Distance Education*, 037(005), 24-34.

Silva, R. , Bernardo, C. D. P. , Watanabe, C. Y. V. , Rosália Maria Passos Da Silva, & José Moreira Da Silva Neto. (2020). Contributions of the internet of things in education as support tool in the educational management decision-making process. *International Journal of Innovation and Learning*, 27(2), 175.

Hosch, B. . (2020). Big data and the transformation of decision making in higher education. *Big data on campus: Data analytics and decision making in higher education*.

Hu Yiling & Gu Xiaoqing. (2019). Assessment Modeling and Empirical Research of Problem- Solving Ability Based on Learning Analytics. *Open Education Research*, 25(02), 107-115.

Avila, C. , Baldiris, S. , Fabregat, R. , & Graf, S. . (2020). Evaluation of a learning analytics tool for supporting teachers in the creation and evaluation of accessible and quality open educational resources. *British Journal of Educational Technology*(3).

Saqr, M. , Nouri, J. , Vartiainen, H. , & Malmberg, J. . (2020). What makes an online problem-based group successful? a learning analytics study using social network analysis. *BMC Medical Education*, 20(80).

Li Yanyan, Zhang Yuan, Su You, Bao Haogang, & Xing Shuang (2019). The Impact of Learning Analytics Tool on Collaborative Learning Performance from the Perspective of Group Awareness. Modern Distance Education Research, 157(01), 106-114.

Garcia-Dias, R. , Vieira, S. , Pinaya, W. H. L. , & Mechelli, A. . (2020). Clustering analysis. *Machine Learning*.

José Maia Neto, Severiano, C. A. , Guimares, F. G. , Castro, C. L. D. , & Cohen, M. W. . (2020). Evolving clustering algorithm based on mixture of typicalities for stream data mining. *Future Generation Computer Systems*, 106.

Brisco, N. D. A. , Nzié Wolfgang, & Serge, D. Y. . (2020). Maintenance modularity optimization using clustering algorithm: application. *International Journal of Industrial Engineering*, 7(1), 12-24.

Liu, S. , & Zou, Y. . (2020). An improved hybrid clustering algorithm based on particle swarm optimization and k-means. *IOP Conference Series Materials ENCE and Engineering*, 750, 012152.

Mulka, M. , & Lorkiewicz, W. . (2020). Different Hierarchical Clustering Methods in Basic-Level Extraction Using Multidendrogram. Information Systems Architecture and Technology: *Proceedings of 40th Anniversary International Conference on Information Systems Architecture and Technology – ISAT 2019*.

Yang Tianpeng & Chen Lifei. (2018). Probability model-based algorithm for non-uniform data clustering. *Journal of Computer Applications*, 38(10), 98-103+283.

Zhao, Y. H. , Shi, H. F. , Ren, X. C. , & Jiao, B. J. . (2020). A toll evasion recognition method based on gaussian mixture clustering. *Communication in Statistics Simulation & Computation*(2), 1-11.

Tao Zhiyong, Liu Xiaofang, & Wang Hezhang. (2018). Clustering algorithm of Gaussian mixture model based on density peaks. *Journal of Computer Applications*, 38(12), 85-89+95.

Martinez-Martin, P. , Jose Manuel Rojo-Abuín, Weintraub, D. , Chaudhuri, K. R. , & Schrag, A. . (2020). Factor analysis and clustering of the movement disorder society-non-motor rating scale: factor analysis and clustering of mds-nms. *Movement Disorder*s(Suppl 1).

Zhang Rong, Chen Yi, Zhang Menglu, & Meng Kexin. (2020). Overviewing of visual analysis approaches for clustering high-dimensional data. *Journal Of Graphics*(1), 44-56.

Wan Jing, Wu fan, He Yunbin, & Li Song. (2020). Clustering Algorithm for High-Dimensional Data Under New Dimensionality Reduction Criteria. *Journal of Frontiers of Computer Science and Technology*(1), 96-107.

Wang Mengyao, Wang Xiaoye, Hong Ruiqi, & Chai Xiaorui. (2019). Opinion Mining Based on BIRCH Clustering Algorithm. *Computer Engineering & Software*, 40(11), 9-12+61.

Zheng Wei, Wang Chaokun, Liu Zhang, & Wang Jianmin. (2010). A Multi-Label Classification Algorithm Based on Random Walk Model. *Chinese Journal of Computers*, 33(8), 1418-1426.

Wu Qiong, Tan Songbo, Xu Hongbo, Duan Juyi, & Cheng Xueqi. (2010). Cross-Domain Opinion Analysis Based on Random-Walk Model. *Journal o f Computer Research and Developm*ent, 047(012), 2123-2131.

Li, K. C. , & Wong, B. T. M. . (2020). The use of student response systems with learning analytics: a review of case studies (2008-2017). *International Journal of Mobile Learning and Organisation*, 14(1), 63.

Cormack, S. H. , Eagle, L. A. , & Davies, M. S. . (2020). A large-scale test of the relationship between procrastination and performance using learning analytics. *Assessment & Evaluation in Higher Education*(1), 1-14.

Sun Hongtao, Li Qiujie, & Zheng Qinhua. (2016). A cluster analysis of MOOC interaction patterns. *Distance Education in China*, 000(003), 33-38,44.

Koh, Y. Y. J. , Schmidt, H. G. , Low-Beer, N. , & Rotgans, J. I. . (2020). Team-based learning analytics: an empirical case study. *Academic Medicine*, 1.

Yarygina, O. . (2020). Learning analytics of CS0 students programming errors: the case of data science minor. *AcademicMindtrek '20: Academic Mindtrek 2020*.

Gosain, A. , & Sachdeva, K. . (2020). Random Walk Grey Wolf Optimizer Algorithm for Materialized View Selection (RWGWOMVS). *Novel Approaches to Information Systems Design*.

Marichamy, V. S. , & Natarajan, V. . (2020). Big Data Performance Analysis on a Hadoop Distributed File System Based on Modified Partitional Clustering Algorithm. *Sustainable Communication Networks and Applicat*ion.

Shi Lingjuan & Huang Decai. (2020). Clustering Algorithm for Position Uncertain Data Expressed by Connection Number. *Journal of Chinese Computer Systems*, 41(2), 361-368.

Liu Jiayao & Wang Jiabin. (2020). Improvement of Slope One Algorithm and Its Implementation on Big Data Platform. *Computer Engineering and Applications*(1), 83-91.

Fan Zhongxin, Wang Xing & Miao Chunsheng. (2019). Improved BIRCH clustering algorithm based on connectivity distance and intensity. *Journal of Computer Applications*, 039(004), 1027-1031.

Izadi, H. , Sadri, J. , Hormozzade, F. , & Fattahpour, V. . (2020). Altered mineral segmentation in thin sections using an incremental-dynamic clustering algorithm. *Engineering Applications of Artificial Intelligence*, 90.

Lucy González, & Lugo, C. . (2020). Fortalecimiento de la práctica docente con learning analytics: estudio de caso.

Wu Fati, Yin Baoyuan & Huang Shihua. (2019). The Research Framework of Learning Habits Dynamics Based on Big Data in Education. *China Educational Technology*, 384(01), 75-81.

Wang Xuenan. (2019). Teachers' Cognition on Big Data in Education and Its Influential Factors: A Survey of 5434 Teachers in China. *Open Education Research*, 25(3): 81-91.

Chen Dexin, Zhan Yuanyuan, & Yang Bing. (2019). Analysis of Applications of Deep Learning in Educational Big Data Mining. *E-education Research*, 040(002), 68-76.

Viloria, A. , Naveda, A. S. , Hugo Hernández Palma, William Niebles Núez, & Leonardo Niebles Núez. (2020). Using big data to determine potential dropouts in higher education. *Journal of Physics Conference Series*, 1432, 012077.

Miah, S. J. , Miah, M. , & Shen, J. . (2020). Editorial note: learning management systems and big data technologies for higher education. *Education and Information Technologies*, 25(2).

Xia Xiaona, Qi Wanxue, Yu Jiguo, & Zou Qi. (2018). The Group Collaborative Behavior of Curriculum under the Perspective of Learning Analytics. *Modern Educational Technology*, 028(009), 47-53.

Jude Osakwe, Gloria E. Iyawa, & Martin Mabeifam Ujakpa (2020). Barriers to the Implementation of Big Data Technology in Education: An Empirical Study. *IST-Africa 2020 Conference Proceedings*, 1-9.

Wu Nanzhong, Huang Zhihu, Zeng Liang, Xie Qingsong, & Xia Haiying. (2019). Construction of education big data ecosystem: logic and practice of "3 + 3" model. *Distance education in China* (7), 77-85

Meng Dongdong, Zhang Yiwei & Cai Yanan. (2020). Big Data's Influence on Higher Education and its Countermeasures. *Heilongjiang Researches on Higher Education*(1), 34-37.

Zhao Yixia, Wang Xin, Jin Kun, & Zhang Peng. (2019). Analysis on the Research Path and Trends of Learning Analysis Technology in Domestic Large Data Environment. *Modern Educational Technology*, 29(8), 34-40.

Bao Haogang, Xing Shuang, Li Yanyan, Zheng Yafeng, & Su You. (2019). Teacher-oriented visual learning analytics tools in online collaborative learning. *Distance education in China*, 000(006), 13-21.